\documentclass{article} 
\usepackage{iclr2024_conference,times}

\usepackage{amsmath,amsfonts,bm}

\def\eqref#1{equation~\ref{#1}}

\def\1{\bm{1}}

\DeclareMathAlphabet{\mathsfit}{\encodingdefault}{\sfdefault}{m}{sl}
\SetMathAlphabet{\mathsfit}{bold}{\encodingdefault}{\sfdefault}{bx}{n}

\usepackage{hyperref}
\usepackage{url}
\usepackage{graphicx}
\usepackage{amsmath}
\usepackage{amssymb}
\usepackage{booktabs}
\usepackage{url}            
\usepackage{amsfonts}       
\usepackage{xcolor}
\usepackage{multirow}
\usepackage{subfig}
\usepackage{newfloat}
\usepackage{listings}
\usepackage{algorithm}
\usepackage{algorithmic}
\usepackage[switch]{lineno}
\usepackage{amsfonts}       
\usepackage{xcolor}
\usepackage{multirow}
\usepackage{subfig}
\usepackage{listings}
\usepackage{newfloat}
\usepackage{wrapfig}

\DeclareCaptionStyle{ruled}{labelfont=normalfont,labelsep=colon,strut=off} 
\floatstyle{ruled}
\newfloat{listing}{tb}{lst}{}
\floatname{listing}{Listing}

\title{Two-Stage Aggregation with Dynamic Local Attention for Irregular Time Series}

\author{Xiaochen Zheng\thanks{The first two authors contribute equally to this work. Correspondence to: Xiaochen Zheng.} \\
Univeristy of Zürich \& ETH AI Center\\
\texttt{xiaochen.zheng@uzh.ch} \\
\And
Xingyu Chen\footnotemark[1]\\
ETH Zürich \\
\texttt{xingyu.chen@yale.edu} \\
\AND
Amina Mollaysa, Manuel Schürch, Ahmed Allam, Michael Krauthammer\\
Univerisy of Zürich \& ETH AI Center\\
\texttt{\{firstname.lastname\}@uzh.ch}
}

\iclrfinalcopy 
\begin{document}

\maketitle

\begin{abstract}
Irregular multivariate time series data is characterized by varying time intervals between consecutive observations of measured variables/signals (i.e., features) and varying sampling rates (i.e., recordings/measurement) across these features. Modeling time series while taking into account these irregularities is still a challenging task for machine learning methods. Here, we introduce TADA, a \textbf{T}wo-stage \textbf{A}ggregation process with \textbf{D}ynamic local \textbf{A}ttention to harmonize time-wise and feature-wise irregularities in multivariate time series. In the first stage, the irregular time series undergoes temporal embedding (TE) using all available features at each time step. This process preserves the contribution of each available feature and generates a fixed-dimensional representation per time step. The second stage introduces a dynamic local attention (DLA) mechanism with adaptive window sizes. DLA aggregates time recordings using feature-specific windows to harmonize irregular time intervals capturing feature-specific sampling rates. Then hierarchical MLP mixer layers process the output of DLA through multiscale patching to leverage information at various scales for the downstream tasks. TADA outperforms state-of-the-art methods on three real-world datasets, including the latest MIMIC IV dataset, and highlights its effectiveness in handling irregular multivariate time series and its potential for various real-world applications. 
\end{abstract}

\section{Introduction}\label{intro}
\begin{wrapfigure}{r}{0.45\textwidth}
  \vspace{-3mm}
   \centering
   \begin{minipage}{0.45\textwidth}
    \centering
    \includegraphics[width=\textwidth]{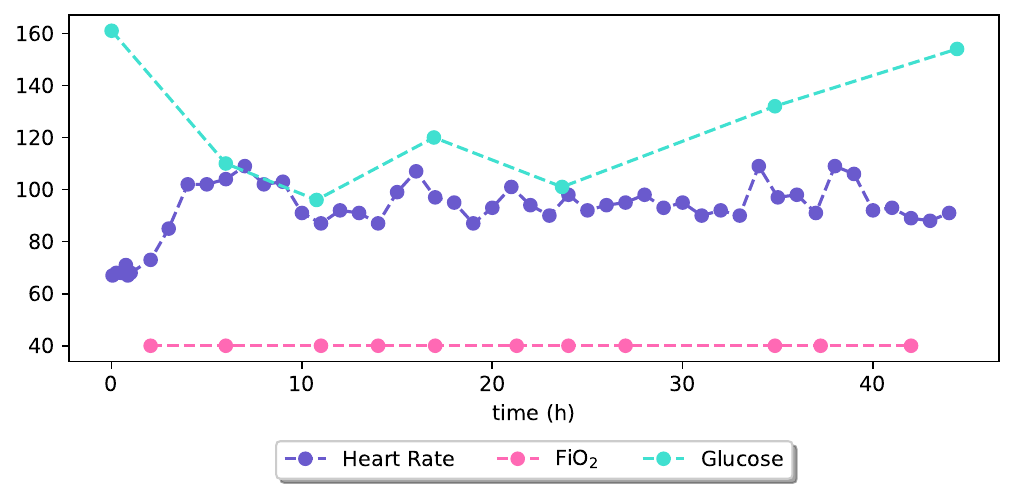}
   \caption{Illustration of selected features in a sample time series from MIMIC IV dataset.}
   \label{fig:realdata}
 \end{minipage}
 \vspace{-3mm}
\end{wrapfigure}

 Irregularly sampled multivariate time series data is prevalent in various domains such as healthcare~\citep{mimiciv,physionet2012}, climate~\citep{braun2022sampling}, economy~\citep{nerlove2014analysis} and environment~\citep{weerakody2021review}. Generally, irregular time series data is characterized by time-wise and feature-wise irregularity. Time-wise irregularity refers to variations in the time intervals (i.e., unequal time spacing) between recorded observations, while feature-wise irregularity refers to the varying sampling rates of the different features (i.e., some features are highly recorded while others are sparsely recorded)~\citep{samplingratereview}. For example, as shown in Figure~\ref{fig:realdata}, each of the three sequences/signals has a varying (i.e., unequal) time intervals between observations indicating time-wise irregularity. Additionally, we can observe that the heart rate signal is sampled more frequently than the glucose signal, highlighting the different sampling rates across features and hence the feature-wise irregularity.

These irregularities in time series data pose challenges for analysis and modeling, as traditional time series methods are based on assumptions of equidistant time intervals and fully observed features at each time step. Over the years, significant progress has been made in adapting machine learning models to better fit the structure of irregular time series. To deal with time-wise irregularities, recent methodologies have proposed new updating equations for RNNs~\citep{GRUD}, employing neural ordinary differential equations (ODEs) to model time dynamics~\citep{latenode}, and suggesting attention mechanisms to convert irregular observations to regularly spaced data~\citep{mtan,ipnet}. These methods capture temporal dependencies across time steps while often having limited ability to learn the correlations between features over time. 
Other methods are proposed to model feature-wise correlations by utilizing graph neural networks~\citep{raindrop} or reformulating observations as unordered set elements~\citep{seft}. However, such methods do not fully consider the multi-scale (i.e., global-local) temporal aspect of the time series information.

In this paper, we propose TADA, a two-stage aggregation process with dynamic local attention followed by hierarchical MLP mixers that account for both time-wise and feature-wise irregularities. Time series aggregation is the process of aggregating multiple feature-wise observations into representative embeddings or aggregating multiple temporal representations into representative time steps~\citep{kotzur2018impact,hoffmann2020review,teichgraeber2022time}. The first stage (feature-wise) aggregation, denoted by temporal embedding, we aim to generate a fixed dimensional representation for each time step using a feature-wise attention mechanism with learned queries. This attention mechanism also learns the weight of each feature in the context of all observations at the same time step. The second stage (time-wise) aggregation aims to harmonize the irregular time intervals through a flexible, dynamic local attention (DLA) mechanism. A key aspect of DLA is the ability to adaptively learn the size of the local attention window specific to each feature by considering that features have varying sampling rates.

Following the learned representations from DLA, we propose stacking multiple MLP-based~\citep{attmlp,mlpreview,mtsmixer} blocks. These blocks learn the multi-scale representations by partitioning the learned representations into patches and merging neighboring patches into a larger scale in the next layer. 

The contributions of our framework are as follows:
\begin{enumerate}
    \item We propose TADA, a two-stage aggregation process with \textbf{dynamic local attention} (DLA) to deal with time-wise and feature-wise irregularities of irregular multivariate time series data. DLA localizes the attention to a learnable range (sliding window) specific to each feature in the time series. Thus effectively addressing the variations in sampling rates across features.
    \item We introduce a Hierarchical MLP Mixer with patching that integrates multi-scale (i.e., local and global) information for classification. Our method outperforms existing models, as demonstrated through comprehensive experiments on three common benchmark datasets. Additionally, we evaluate its performance using the latest MIMIC-IV~\citep{mimiciv} dataset, providing processing and experiment code benchmarks for future research.
\end{enumerate}

\section{Related Work}\label{related}
Irregular time series is characterized by its uneven time intervals between observations. In the multivariate setting, it also has variable sampling rates and a lack of alignment across different features. This presents a unique challenge as traditional methods for time series modeling typically require uniformly spaced and aligned inputs~\citep{tsregular}.

To overcome these obstacles, one approach is to adapt traditional recurrent neural networks to integrate irregular time series by modifying parts of the model architectures. For instance, PhasedLSTM~\citep{phasedLSTM} enhanced the LSTM~\citep{LSTM} model by incorporating a novel time gate (i.e. computational unit) accounting for irregular time intervals. Likewise, the authors in \cite{GRUD} developed several versions of the Gated Recurrent Unit (GRU) to include masking and time intervals directly into the GRU architecture. They proposed GRU-D, which introduces an additional gate to decay the last observed values selectively. More recent methodologies~\citep{latenode,ode} used ordinary differential equations (ODE) in neural networks to learn latent states of an ODE from the observed time series data, effectively capturing underlying dynamics and dependencies. While these methods consider irregular time intervals,  they fall short in exploring the relationships among features and their varying sampling rates.

Another strategy to tackle these challenges is through interpolating the irregular observations into uniform dimensions. IPNet~\citep{ipnet} and mTAN~\citep{mtan} introduced fixed and learnable time-based attention, respectively, to transform input data into a set of regular sampled points. However, these methods are limited as they overlook various scales of information (i.e., global/local information) and apply the same techniques to features with different sampling rates. Other techniques that avoid interpolation such as SeFT~\citep{seft}, model irregular observations as an unordered set of elements and employ neural set functions for classification. However, this method is not suited for handling long sequences as the aggregation function tends to lose information (i.e. discard history) when dealing with large set cardinalities. In contrast, Raindrop~\citep{raindrop} utilizes graph neural networks to capture correlations between features, but still overlooks multi-scale information, which is crucial for time series classification~\citep{multiscale}. A recent study proposing Warpformer~\citep{warpformer} utilized dynamic time warping to handle irregular lengths across various temporal signals and incorporated an attention mechanism to understand the correlations between different features. 

In this work, we identify two approaches that are relevant: (1) the first involves interpolating irregular measurements/observations ~\citep{mtan,ipnet}, and (2) the second focuses on transforming time series into different scales, as explored in \cite{warpformer}. However, both approaches have their disadvantages. The former overlooks the different sampling rates across different features in irregular time series, and the latter relies on dynamic time warping to learn information at each scale, which requires high computational complexity and hyperparameters specific to each scale. In contrast, our proposed method considers both the temporal and feature-wise irregularities when unifying irregular time series and introduces an adaptive attention method that alleviates the need for pre-defined hyperparameters.
\section{Methods}
\begin{figure*}[!t]
  \centering
    \subfloat[\scriptsize Irregular Representation]{
    \label{fig:data repr}
    \includegraphics[width=0.2\linewidth]{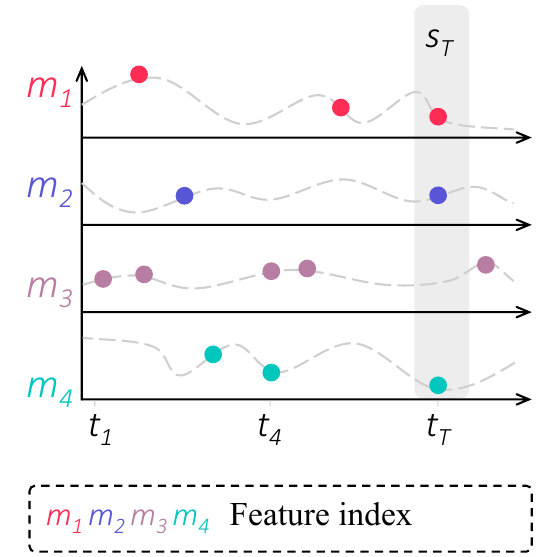}}
   \subfloat[\scriptsize Temporal Embedding Module Overview]{
    \label{fig:embedding}
    \includegraphics[width=0.59\linewidth]{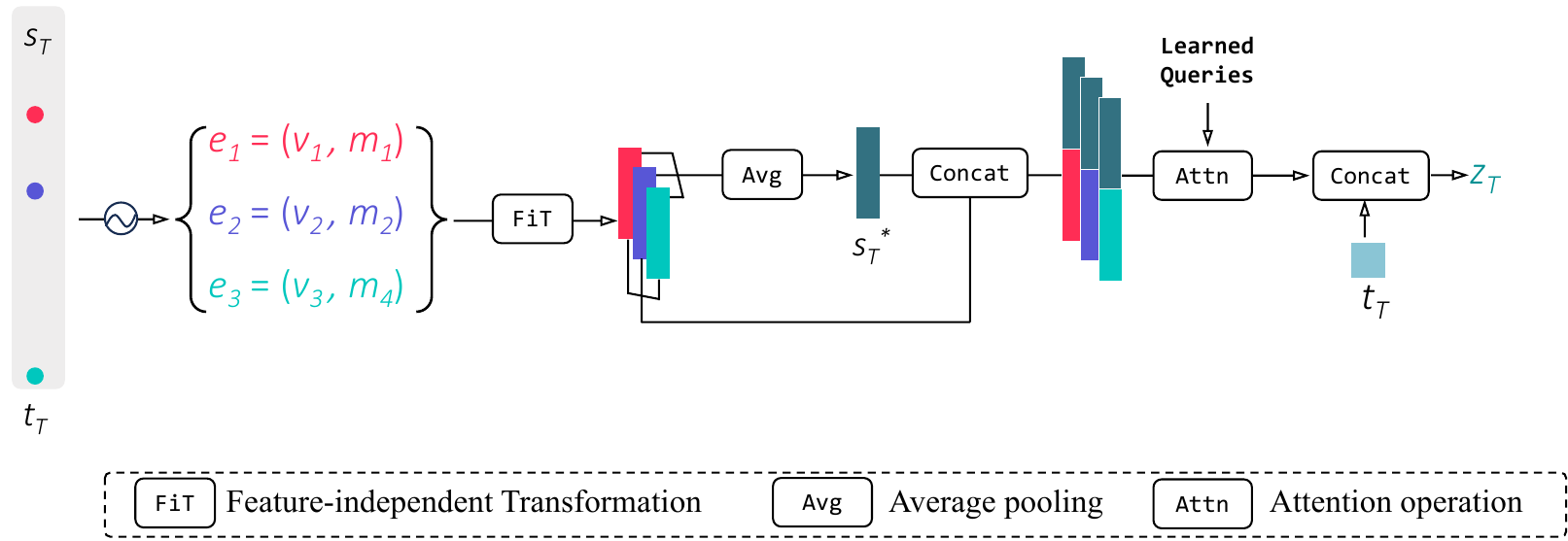}}
    
  \caption{\label{fig:repr} Data Structure and Temporal Embedding Module
  }
\end{figure*}

\subsection{Problem Definition}
\label{notation}
Consider a time series denoted as $x$, which extends over $T$ time steps and contains up to $D$ features. At a particular time $t_k$, we assume that the number of observed features is $n_k$, where $n_k \leq D$ and $k \leq T$. We represent the observed features at $k$-th time step as a set of pairs $s_k = \left \{e_j\right \}_{j=1}^{n_k} = \left \{(v_j, m_j) \right\}_{j=1}^{n_k}$, which includes the observed value $v_j \in \mathbb{R}$ and the corresponding feature index $m_j \in \left \{1,...,D\right \}$. 

We denote the time series $x$ as the sequence of tuples $(s_k, t_k)$:
\begin{equation}
    x = \{(s_k, t_k)\}_{k=1}^{T}
\end{equation}
where $s_k$ are the observed features at time $t_k$ and $T$ represents the length of the sequence. We illustrate our data representation in Figure~\ref{fig:embedding}. Note that at each time step, we expect to have different sizes of vectors due to unobserved features. We include any time step into the representations with at least one observed feature. Furthermore, we effectively handle irregular time intervals between observations by explicitly including time values $t_k$. 

\begin{figure*}[ht!]
    \includegraphics[width=0.9\textwidth]{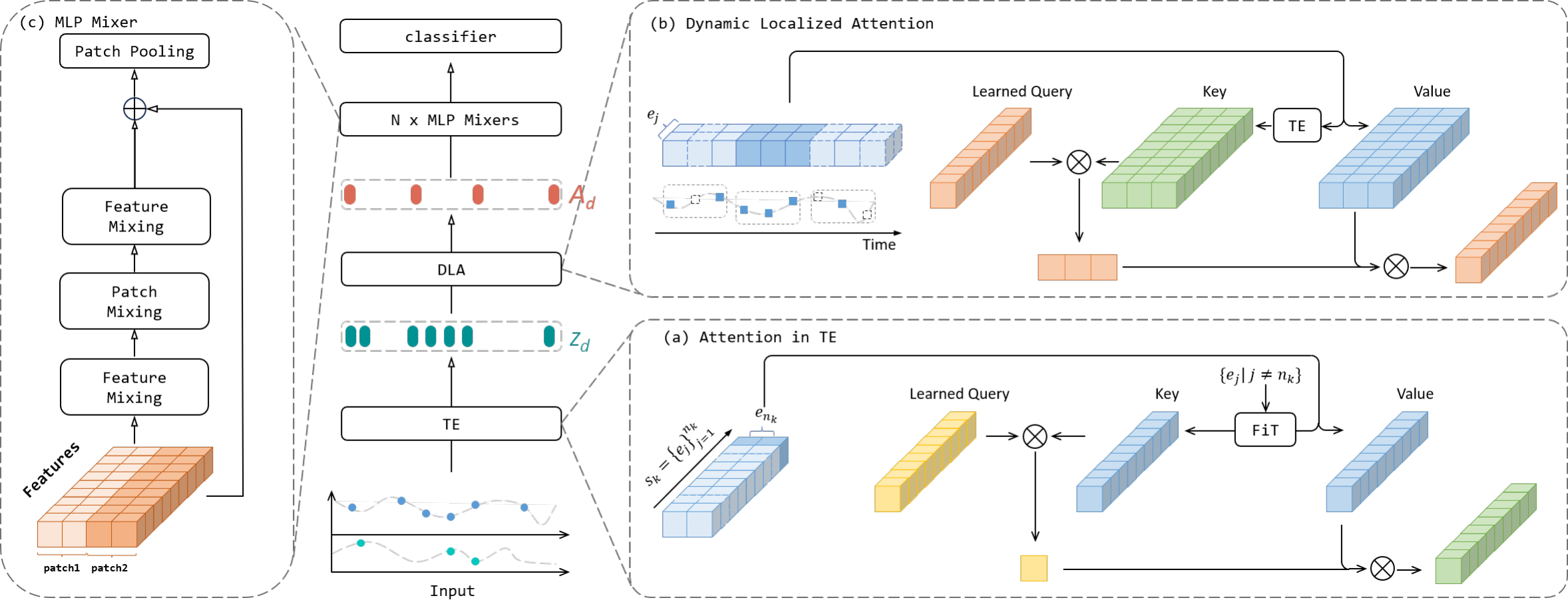}
    \centering
    \caption{Overview of TADA architecture. (a) shows the temporal embedding (TE) at a specific time step over all features. (b) shows the dynamic localized attention on a specific feature across all observed time steps. (c) shows the MLP mixers accepting learned representations from DLA as inputs. The activation and normalization layer is omitted for simplicity. The figure shows a segmentation of the representations using a patch size of two.}
    \label{fig:model}
\end{figure*}
\subsection{Two-Stage Aggregation}~\label{sec:Two stage}
\subsubsection{Stage 1: Temporal Embedding Module (\texttt{TE})}\label{sec:temporal embed} In irregular time series, different subsets of features are present at each observed time step. To deal with the different sizes of subsets, the most common way is to use aggregation methods, such as mean and average, to encode them into a fixed-dimensional embedding. However, different features have different importance in determining its final classification results~\citep{raindrop,seft}. Therefore, we design the first stage of aggregation of TADA, the temporal embedding module (\texttt{TE}), which focuses on learning a fixed dimensional embedded representation for each time step. The first stage learns to aggregate all available features at each time step while capturing individual features' importance in the learned representation. 

To achieve these two goals, we employ scaled dot-product attention to learn embeddings for an arbitrary number of observations at each time step. The overview of the Temporal Embedding Module is displayed in Figure~\ref{fig:embedding}. Specifically, for each time step, we first learn the embedding of each observed feature at time $t_k$ using a Feature-independent Transformation (\texttt{FiT}) function: $\texttt{FiT}(e_j): \mathbb{R}^2\rightarrow \mathbb{R}^{d_g}$. The \texttt{FiT} can be any linear or non-linear transformation function. By aggregating the learned representation from all the observed features, we obtain a fixed dimensional feature representation $s^*_k$ for the given time step $t_k$:
\begin{equation}\label{aggre}
    s^*_k = \frac{1}{n}\sum_{j=1}^{n}\texttt{FiT}(e_j)
\end{equation} 
where $s_k^*\in R^{d_g}$ and it can be seen as a summary statistic of all observed features at time $t_k$~\citep{deepset}.

To account for the individual feature significance at each time step, we suggest an attention mechanism to aggregate features:
\begin{align}\label{eq:embed_K}
    K_j &=  U_k\text{concat}\left[s^*_k, e_j\right] \\
    \texttt{attn}_k &= \sum_{j=1}^{n_k}\frac{\text{exp}(\left \langle u,K_j \right \rangle/\sqrt{d_{\text{embed}}})}{\sum_{i=1}^{n_k}\text{exp}(\left \langle u,K_i \right \rangle/\sqrt{d_{\text{embed}}})}U_ve_{j}\label{eq:temporal attn}
\end{align}
where $U_k\in\mathbb{R}^{(d_g+2)\times d_{\text{embed}}}, U_v\in\mathbb{R}^{2\times d_{\text{embed}}}$ are projection matrices, Equation \ref{eq:temporal attn} is the scaled dot-product attention computation. As seen from equation~\ref{eq:embed_K}, $K_j\in\mathbb{R}^{1\times d_{\text{embed}}}$ represents the key vector for element $e_j$. 

To construct the value vector, we apply $U_v$ to project $e_j$ and initialize a learnable query vector $u\in\mathbb{R}^{1\times d_{\text{embed}}}$ for all features.

Subsequently, the model progressively learns to assign weights to different features and generate the sum of a weighted representation of observed features for each time step. Then we concatenate the time $t_k$ to the resulting vector to get $z$ $\in \mathbb{R}^{d_{\text{embed}}+1}$, as shown in equation~\ref{eq:score}. 
\begin{equation}\label{eq:score}
     z_k = \text{concat}\left[t_k, \texttt{attn}_k\right]
\end{equation}
Horn \textit{et al.}~\citep{seft} proposes SeFT, which treats every observation at every available time step as an individual set element and aggregates all sets by an attention mechanism, considering relationships between elements and learning an embedding insensitive to the number of available features. However, with a large number of observations, the numerical imprecision of aggregation methods may cause small attention values to be disregarded. Contrary to handling all available observations simultaneously, we first group the observations from the same time step and then independently learn the embeddings of the observation for each time step. In this way, we avoid losing information when dealing with a large number of temporal observations.

\subsubsection{Stage 2: Dynamic Local Attention (\texttt{DLA})}
In the first stage, the \texttt{TE} module allows us to compute an embedding for each time step by learning different weights for features within each time step. However, \texttt{TE} only processes individual time steps and does not explicitly consider uneven time intervals and different sampling rates across features. This limitation of \texttt{TE} necessitates the second stage of our method: Dynamic Local Attention (\texttt{DLA}). It is a localized attention mechanism that considers the different sampling rates of each feature and uneven time intervals.

The proposed TE module returns a fixed dimensional representation $z_k = \texttt{TE}(s_k) \in \mathbb{R}^{d_{\text{embed}}+1}$ for each time step $t_k$ of the irregular time series $x$. The transformed time series can be represented as $\hat{x} = \left \{ z_1, z_2, ...,z_T \right \} \in \mathbb{R}^{T\times (d_{\text{embed}}+1)}$, where $T$ denotes the length of $x$. Similar to scaled-dot product attention, \texttt{DLA} requires key, value, and query pairs as input. We use $\hat{x}$ as the input key matrix, and we construct $V \in \mathbb{R}^{T\times D}$ by applying the masking function proposed in~\citep{maskingfunction} to the raw signals to represent the value matrix. For the query matrix, we initialize $Q \in \mathbb{R}^{L\times(d_{\text{embed}}+1)}$, where $L$ is the total number of queries and each row is a learnable individual query vector used for aggregating information from keys through the attention computation.

Given a query $Q$ and a key $\hat{x}$, the local attention for a single head is defined as follows:
\begin{equation}\label{eq:dat}
    A_{id} = \texttt{DLA}(Q,\hat{x},V)_{id} = \sum_{j=0}^{T}\frac{\text{exp}(\alpha_{ijd})}{\sum_{j'\in \mathcal{R}(i,d)}\text{exp}(\alpha_{ij'd})}v_{jd}
\end{equation} where
\begin{equation}\label{eq:alpha}
    \alpha_{ijd} = \left \langle W_qQ_i,W_k\hat{x}_j \right \rangle /\sqrt{d_{\text{proj}}}
\end{equation}
In Eq.\ref{eq:alpha}, $Q_i$ and $\hat{x}_j$ are the $i$-th query in $Q$ and the $j$-th time steps of $\hat{x}$. $W_q,W_k\in\mathbb{R}^{(d_{\text{embed}}+1)\times d_{\text{proj}}}$ are linear projections for query and keys. Eq.\ref{eq:alpha} calculates the attention weights $\alpha_{ijd}$ from the scaled dot-product between the $i$-th query and the $j$-th time point on feature $d$. 

It is noted that to learn a local attention window specific to each feature, we introduce a range function $\mathcal{R}$ in the denominator of Equation~\ref{eq:dat}. For a given feature dimension $d$, we define the local neighborhood of the $i$-th query as a collection comprising time points in close proximity within the range of $r_d$. The center of the neighborhood for the $i$-th query is defined as a regular anchor time, $t_i^q=i\frac{t_T}{L}$, given the maximum observed time $t_T$ and $L$ initialized queries. To be more specific:
\begin{equation}\label{eq:r}
\begin{aligned}
   \mathcal{R}(i,d) &= \{ j'|t_{j'}\in [max(0,t_i^q-r_d),min(t_T,t_i^q+r_d)]\}\\
   &\textit{where} \:\:\:1\leq j'\leq T, \textit{and}\:\:j'\in\mathbb{N}
\end{aligned}
\end{equation}
where $r_d$ is the learnable attention window size specific to $d$-th feature. In Equation~\ref{eq:r}, for feature $d$, each individual query $i$ aggregates values $v_{j'}$ if its observed time $t_{j'}$ is within a time span of $r_d$ around the anchor time $t_i^q$. In short, the range function generates varying sizes of local neighborhoods for individual features, accounting for their respective sampling rates.

In Eq.\ref{eq:dat}, $A_{id}$ is aggregated information from the temporal neighborhood of $d$-th feature around $i$-th query. We can obtain $A\in \mathbb{R}^{L \times D}$ containing aggregated information from $D$ features with corresponding $L$ queries. And we further apply a linear layer to project $A$ to $\mathbb{R}^{L\times d_{\text{patch}}}$ as the final output of \texttt{DLA}. The multi-head attention is an extension of \texttt{DLA}, which runs several \texttt{DLA} in parallel. Assuming we have $p$ heads in total, the final output representation is $\text{concat}(A^1,..., A^p)$.

The key distinction between the common scaled-dot product attention and our approach is that instead of computing attention scores overall observed time steps, we restrict the attention range to a neighborhood of time steps around each query vector. Moreover, to account for the different sampling rates, we learn the range of the neighborhood for each dimension through a range function. The output attention score, therefore, includes the feature sampling rate information and can assign a different attention neighborhood to each feature. 

\subsection{Hierarchical MLP Mixer}
With the two-stage attention-based aggregation, we have transformed the irregular time series into a learned representation $A\in\mathbb{R}^{L\times d_{\text{patch}}}$ by addressing both feature and time irregularity. To better explore the global and local patterns exhibited in the time series, we introduce a hierarchical MLP-based architecture.  Recent studies in computer vision have shown that MLP-based models can adeptly learn patterns by iteratively combining information across patches and channels. It allows communications with different channels and locations (patches)~\citep{mtsmixer}. Furthermore, the hierarchical structure is proved effective in learning different scales of information for time series~\citep{crossformer,warpformer,fedformer}. 

To effectively explore both local and global patterns in the original time series, we present a hierarchical structure using multiple stacked MLP Mixer blocks. Each of these MLP Mixers operates in two distinct mixing directions: \textit{patch mixing} and \textit{feature mixing}. The initial input for the first MLP mixer is formed by segmenting the output from \texttt{DLA}, represented as $A\in \mathbb{R}^{L\times d_{\textbf{patch}}}$, into non-overlapping patches with length $p$.  The patching process generates a sequence of patches $\beta^0_{i,:}\in\mathbb{R}^{p\times d_\text{patch}}$, where $1\leq i\leq L/p$.  For convenience, we concatenate all patches together as $\beta^0 = \text{concat}[\beta^0_{i,:},\beta^0_{2,:},...\beta^0_{L_0,:}]$, where $\beta^0 \in \mathbb{R}^{L_0 \times p \times d_{\text{patch}}}$, $L_0 = L/p$. 

The overview of each layer is illustrated in block (c) of Figure~\ref{fig:model}. The concatenated patches are inputs for each MLP mixer block. For $l\geq 1$, each block is defined by: 
\begin{equation}\label{eq: mlpblock}
\begin{gathered}
    \hat{\beta}^{l} = \sigma(\beta^{l-1}+(W_2\beta^{l-1}W_1)W_3)
\end{gathered}
\end{equation}
where $\sigma$ denotes the nonlinear activation functions such as ReLU~\citep{relu}. Normalization is omitted for simplicity. The matrices $W_1$ and $W_3$, both of dimensions $d_{\text{patch}} \times d_{\text{patch}}$, function as linear projections operating across the feature dimensions (i.e., \textit{feature mixing}). $W_2\in\mathbb{R}^{L_{l-1}\times L_{l-1}}$ is a matrix for which the size $L_{l-1}$ corresponds to the number of patches at layer $l-1$. It performs the linear operation along the first axis of $\beta^{l-1}$ to capture inter-patch interactions (i.e., \textit{patch mixing}). For example, if the sequence has five patches, the shape for $W_2$ is $5\times 5$. 

The \textit{feature mixing} and \textit{patch mixing} capture information on the corresponding scale in block $l$. Then we merge every $m$ adjacent patch to generate a coraser representation for the next block:
\begin{equation}\label{eq:patch merge}
\begin{aligned}
    \beta^{l}_{i,:} &= \text{concat}\left [ W_4 \hat{\beta}^{l}_{mi-1,:},...,W_4 \hat{\beta}^{l}_{mi-(m-1),:} \right ]
\end{aligned}
\end{equation}
where $\dot{\beta}_{i,:}^l \in\mathbb{R}^{p\times d_{\text{patch}}}$ denotes the $i$th patch in $\dot{\beta}^l$, constrained by $1\leq i\leq \frac{L_{l}}{p/m}$. Each patch of length $p$ is transformed to a length of $p/m$ through $W_4\in\mathbb{R}^{p\times p/m}$. 

To utilize the information present at various representation scales, we pass the output of each layer's MLP mixers through a fusion block. In the fusion block, we apply an average pooling layer to the output of each layer, resulting in representations of consistent length. These representations are then combined through element-wise multiplication and an MLP layer. The final model output is generated via an MLP classifier. The detailed structures of the MLP mixers, fusion methods, and classifier are in the appendix.

\section{Experiments}

\subsection{Baselines}
We benchmark our model against recent state-of-the-art methods for irregular multivariate time series. \textbf{GRU-Simple}~\citep{GRUD} handles missing variables and variable time intervals by utilizing the measurement, masking, and time interval vectors as inputs to a Gated Recurrent Unit (GRU) network. \textbf{GRU-D}~\citep{GRUD} is based on GRUs and integrates a decay mechanism in its internal state to account for irregular time intervals. \textbf{Latent-ODE} and \textbf{ODE-RNN}~\citep{latenode} leverage neural ODEs to capture continuous dynamics from the data. The \textbf{mTAND}~\citep{mtan} uses continuous time attention to interpolate irregular time intervals into a fixed-dimensional space. \textbf{SeFT}~\citep{seft} and \textbf{RainDrop}~\citep{raindrop} take different approaches by reformulating irregular time series as unordered sets or graphs, respectively. \textbf{Warpformer}~\citep{warpformer} uses dynamic time warping and a transformer-like structure to capture multi-scale information from irregular time series. We provide further details on dataset used in Appendix.

\subsection{Implementation Details}
All methods were trained using the Adam optimizer, and their hyperparameters were selected based on the validation sets. To ensure fair comparison with other baselines, we utilized the best hyperparameters reported for the baseline methods. Each dataset was randomly split using a specified random seed, which was also employed to initialize the models. We repeated the experiments five times, each with a different random seed. The reported results represent the average performance across the five experiments.

We performed classification experiments on all three datasets: PhysioNet and MIMIC IV datasets have one label for each time series, while the human activity dataset has a label for each time point in the time series. Due to the class imbalance in the PhysioNet and MIMIC IV datasets, we evaluated the model performance using the area under the ROC curve (AUROC) and the area under the Precision-Recall Curve (AUPRC). Since the Human Activity is a balanced dataset, and to ensure a fair comparison to other reported methods, we used accuracy and AUPRC given that the main baselines such as mTAND~\citep{mtan} and Warpformer~\citep{warpformer} used accuracy as a performance metric for this dataset.
\begin{table*}[t]
\centering
\footnotesize
    \caption{Classification Performance on PhysioNet, MIMIC IV and Human Activity dataset}
    \label{table:main}
    \resizebox{\textwidth}{!}{
     \begin{tabular}[h]{lcccccc}
        \toprule
        \multirow{2}{*}{\bf Model} & \multicolumn{2}{c}{\bf PhysioNet} & \multicolumn{2}{c}{\bf MIMIC IV}& \multicolumn{2}{c}{\bf Human Activity} \\
        \cmidrule{2-3}
        \cmidrule{4-7}
        & {\bf AUROC} & {\bf AUPRC} & {\bf AUROC} & {\bf AUPRC} & {\bf Accuracy} & {\bf AUPRC}\\
        \midrule
        GRU-Simple	    & $	0.737	\pm	0.010	$ & $0.335 \pm 0.044$ & $	0.681	\pm	0.005	$ & $ 0.083 \pm 0.002$& $	0.774\pm0.014	$ & $0.661\pm0.020$ \\
        GRU-Decay	    & $	0.787	\pm	0.014	$ & $0.389 \pm 0.020$ & $	0.758	\pm	0.011	$ & $ 0.238 \pm 0.016$& $	0.749	\pm	0.014	$ & $0.655 \pm 0.024$ \\
        Latent-ODE     & $ 0.837   \pm 0.006   $ & $0.533 \pm 0.011$ & $ 0.763   \pm 0.010   $ & $ 0.248   \pm 0.015   $ & $ 0.767   \pm 0.014   $ & $0.649 \pm 0.026$  \\     
        ODE-RNN	        & $	0.812	\pm	0.013	$ & $0.466 \pm 0.036	$ & $0.780 \pm 0.017$ & $0.301 \pm 0.029$& $	0.767	\pm	0.014	$ & $0.771 \pm 0.014$\\
        SeFT	        & $	0.733	\pm	0.210	$ & $0.313 \pm 0.038$ & $ 0.783   \pm 0.012$ & $ 0.277 \pm 0.023 $& $	0.758	\pm	0.033	$ & $0.661 \pm 0.033$\\
        mTAND	& $	0.848	\pm	0.090	$ & $0.514 \pm 0.028$ & $	0.738 \pm 0.010 $ & $ 0.226 \pm 0.030$& $	0.813	\pm	0.030	$ & $0.735 \pm 0.080$ \\
        RainDrop	& $	0.711	\pm	0.017	$ & $0.258 \pm 0.022$ & $	0.781	\pm	0.010	$ & $	0.269	\pm	0.020	$& $	0.702	\pm	0.016	$ & $0.623 \pm 0.026$\\
        Warpformer	& $	0.855	\pm	0.005	$ & $0.534 \pm 0.019$ & $0.804\pm 0.017 $ & $ 0.317 \pm 0.035$&$0.849\pm 0.070$ & $0.811\pm 0.900$\\
        \midrule 
        
        {Ours}& $	\bf{0.862	\pm	0.007}	$ & $\bf{0.551} \pm \bf{0.028}$ & $   \bf{0.817 \pm 0.006}$ & $ \bf{0.335 \pm 0.023}$&$\bf{0.907	\pm	0.004}	$ & $\bf{0.906} \pm \bf{0.007}$ \\
        \bottomrule
    \end{tabular}
    }
\end{table*}

\section{Results}

Table~\ref{table:main} compares the performance of all baseline methods on whole time-series classification in PhysioNet, MIMIC IV and Human Activity datasets. Overall, our approach consistently surpasses the baselines across all datasets, with a noticeable difference in MIMIC IV and Human Activity dataset, highlighting the capability of our approach to excel under diverse scenarios. 
Among baselines, we noticed that SeFT performs well on MIMIC IV and Human Activity dataset, but it does not exhibit comparable performance on the Physionet dataset. We hypothesize that the underlying reason is related to the aggregation operations on set elements that might lead to loss of information when the set cardinality is large. Warpformer and mTAND excel on the PhysioNet and Human Activity datasets, yet fall short on the MIMIC IV dataset. Furthermore, the Raindrop method, although relatively effective on the MIMIC IV dataset, is even unable to surpass GRU-Simple on the PhysioNet dataset. This could be due to Raindrop's limitations in modeling dimension correlations effectively when the dimensionality is high. 

\subsection{Hyperparameters}
In this section, we analyze the effect of different hyperparameters on our model performance. We begin by investigating the influence of the number of queries $L$ in \texttt{DLA} and the patch size $p$ on our model's performance. The first layer of MLP mixer takes input with length $L$ and partitions it into patches with size $p$. 

\begin{figure}[!t]
  \centering
    \subfloat[Human Activity]{
    \label{fig: act_grid_patch1}
    \includegraphics[width=0.3\linewidth]{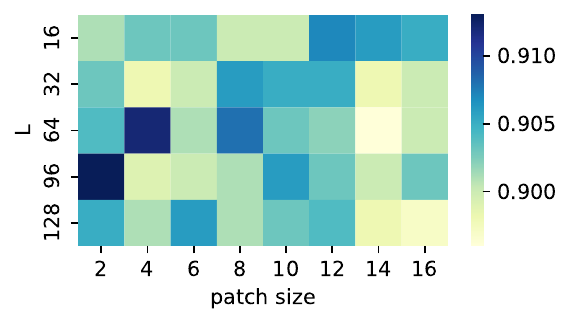}}
   \subfloat[PhysioNet]{
    \label{fig:act_grid_patch1}
    \includegraphics[width=0.3\linewidth]{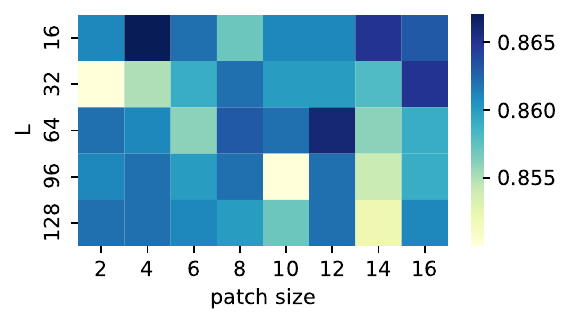}}\\
    \subfloat[Human Activity]{
    \label{fig: act_grid_patch2}
    \includegraphics[width=0.3\linewidth]{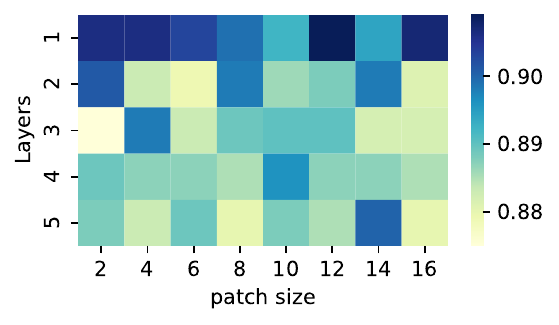}}
   \subfloat[PhysioNet]{
    \label{fig:phy_grid_patch}
    \includegraphics[width=0.3\linewidth]{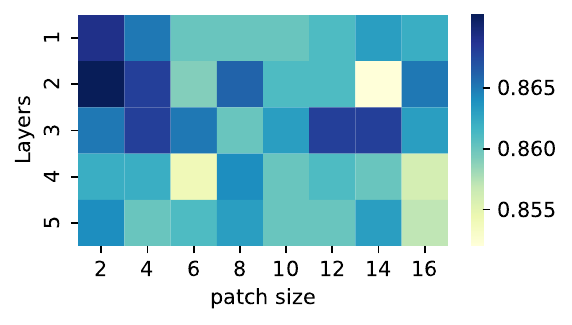}}
  \caption{\label{fig:grid_patch}  Evaluation on hyper-parameter influence. The model performances corresponding to (a), (b) different combinations of patch sizes and number of attention queries ($L$) , to (c) (d) different combinations of patch sizes and MLP mixer layers for the Human Activity and PhysioNet datasets.
  }
\end{figure}

In figure~\ref{fig:grid_patch}, we fix the number of layers while changing the patch size and the number of queries on the Human Activity and PhysioNet datasets. For the Human Activity dataset, we observe that a higher number of queries and smaller patch sizes tend to yield optimal performance. This result is likely because we perform classification at each time step for Human Activity dataset, which needs detailed information. Conversely, in the PhysioNet dataset, we find that a lower number of queries and larger patch sizes tend to achieve optimal performance. This is because the PhysioNet dataset only requires predicting a single label for all time steps within a sample, making coarser scale information sufficient for this task. We also investigated the impact of different combinations of patch length and layers while fixing the number of queries. These two hyperparameters determine the number and size of scales in MLP mixers. The optimal number of layers and patch size vary across different datasets. In particular, small patch sizes and 1-2 layers are sufficient to produce good results. 

\begin{table*}[t!]
\centering
\caption{Ablation tests of TADA on PhysioNet, Human Activity and MIMIC IV dataset}
\resizebox{\textwidth}{!}{
\begin{tabular}{lcccccc}
\toprule
\multirow{2}{*}{\textbf{Model}} & \multicolumn{2}{c}{\textbf{PhysioNet}} & \multicolumn{2}{c}{\textbf{MIMIC IV}} & \multicolumn{2}{c}{\textbf{Human Activity}}  \\
 & AUROC & AUPRC & Accuracy & AUPRC& AUROC & AUPRC \\
\midrule
Full & $	\bf{0.862	\pm	0.007}	$ & $\bf{0.551} \pm \bf{0.028}$ & $   \bf{0.817 \pm 0.006}$ & $ \bf{0.335 \pm 0.023}$&$\bf{0.907	\pm	0.004}	$ & $\bf{0.906} \pm \bf{0.007}$ \\
\midrule
w/o \texttt{DLA} & 0.499 ± 0.002 & 0.144 ± 0.007 & 0.490 ± 0.001 & 0.087 ± 0.000& 0.380 ± 0.007 & 0.143 ± 0.001\\
w/o learnable range& 0.859 ± 0.009 & 0.541 ± 0.019& 0.788 ± 0.004 & 0.301 ± 0.012 & 0.892 ± 0.009 & 0.851 ± 0.009 \\
w/o MLP mixer & 0.855 ± 0.007 & 0.549 ± 0.031 & 0.770 ± 0.002 & 0.256 ± 0.067 & 0.756 ± 0.012 & 0.638 ± 0.025\\
\bottomrule
\end{tabular}
}
\label{tab:ablation}
\vspace{-0.1in}
\end{table*}

\subsection{Attention Analysis}

\begin{figure}[ht!]
  \centering
    \subfloat[Diastolic Blood Pressure]{
    \label{fig:attn_dbp}
    \includegraphics[width=0.25\linewidth]{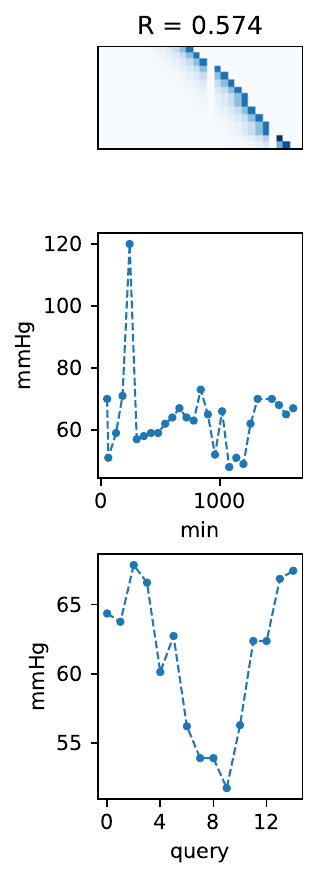}}
   \subfloat[Temperature]{
    \label{fig:attn_temp}
    \includegraphics[width=0.25\linewidth]{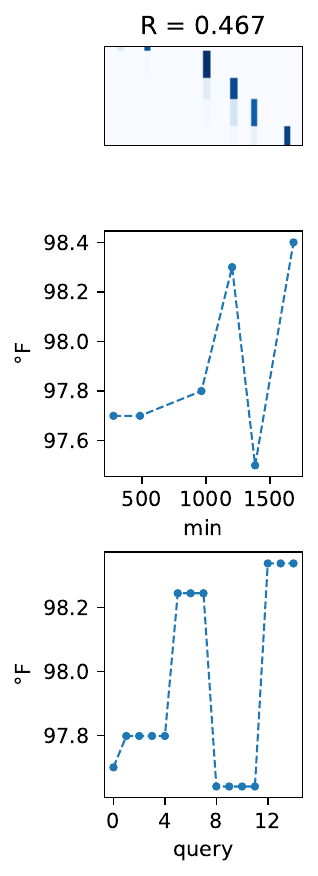}}
    \subfloat[SpO$_2$]{
    \label{fig:attn_spo2}
    \includegraphics[width=0.25\linewidth]{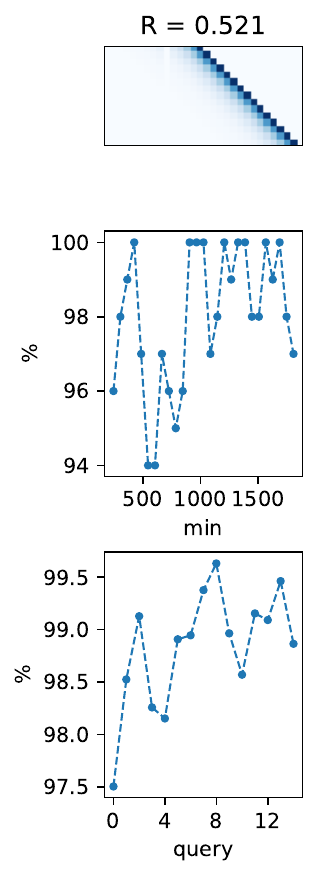}}
    \subfloat[Glucose]{
    \label{fig:attn_glucose}
    \includegraphics[width=0.25\linewidth]{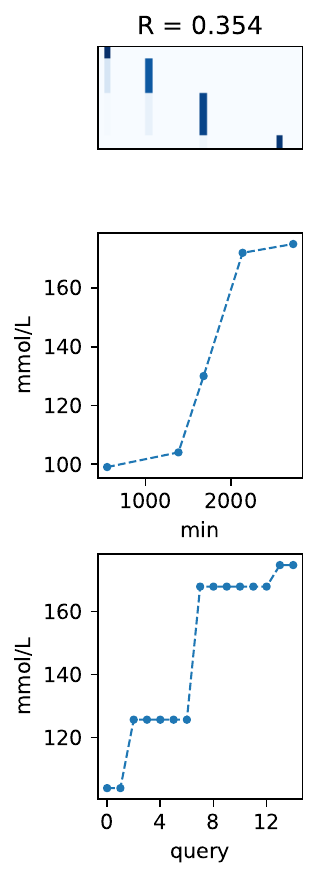}}
  \caption{DLA visualization on feature (a) Diastolic Blood Pressure, (b) Temperature, (c) SpO$_2$ and (d) Glucose in MIMIC IV dataset. R denotes the learned window size in their respective attention scores (after normalization). Top row: Attention score of \texttt{DLA}. Middle row: Original values of signals. Bottom row: Weighted queries after \texttt{DLA}.} \label{fig:attn}
\end{figure}

To further illustrate the attention mechanism of \texttt{DLA} and its handling of different sampling rates across features, we present attention scores from samples of the MIMIC IV dataset and the signals before and after \texttt{DLA} module in Figure~\ref{fig:attn}. The figure reveals \texttt{DLA} is able to learn different window sizes for different features. For example, diastolic blood pressure and SpO$_2$ have larger attention window sizes to capture a general trend. Enlarging the window also aids in removing outlier values, as depicted in Figure~\ref{fig:attn_dbp}. On the other hand, with a smaller window size, as demonstrated in Figure~\ref{fig:attn_temp} and \ref{fig:attn_glucose}, the output queries successfully upsample observations while still capturing the overall trend. This analysis highlights the adaptive nature of our \texttt{DLA} module that is capable of dynamically adjusting the attention window based on varying sampling rates for each feature. This adaptive capability not only provides more flexibility but also enhances accuracy when unifying irregularly spaced observations.

\subsection{Ablation Studies}
We carried out ablation studies on TADA to demonstrate the significance of its modules. Table~\ref{tab:ablation} presents the results of different variants. In the first set of experiments, we focused on the dynamic local attention module (\texttt{DLA}). Specifically, we conducted two separate experiments by either removing the entire \texttt{DLA} block (\textit{`w/o \texttt{DLA}'}) or dropping the learnable range part individually (\textit{`w/o learnable range'}) and have the attention computed over all time steps without localized windows. The findings suggest that the \texttt{DLA} module plays a crucial role in our classification tasks, as its absence leads to a drastic decrease in performance. Additionally, the learnable attention range provided the \texttt{DLA} with enhanced flexibility across various datasets, as the absence of a learnable range leads to inferior performance, especially on MIMIC IV  and Human Activity datasets. These results emphasize the importance of simultaneously unifying time intervals and capturing different feature sampling rates for irregular multivariate time series.

It is worth noting that completely removing \texttt{DLA} is equivalent to directly applying the MLP Mixer on feature embeddings. The reduction in performance when using the MLP Mixer alone suggests that it is ineffective in learning patterns for irregular time series. We also investigated the effect of hierarchical MLP mixer blocks. The terms \textit{`w/o MLP mixers'} refers to the removal of the MLP mixer block. The results demonstrate the absence of a hierarchical structure will lead to a performance decrease. Particularly, the absence of the MLP mixer block has the most pronounced impact on the performance on the Human Activity dataset and the least impact on the Physionet dataset. This is intuitive since Human Activity involves classification at each time step, thus demanding more detailed information compared to the other two datasets. Lastly, the \texttt{DLA} module alone seems to be sufficiently capable of extracting meaningful patterns from the Physionet dataset with the highest number of features.

Additionally, we explore how various fusion techniques applied to the outputs of the different layers can influence the overall performance of the model. We investigated the effect of merging outputs from distinct scales through different methods such as element-wise multiplication, addition, and concatenation. As shown in Table~\ref{tab:fusion}, the differences in performance resulting from these fusion methods are relatively subtle. However, it is noteworthy that among the three methods employed, multiplication yields the most optimal performances. This suggests that while the impact might not be significantly pronounced, employing a multiplication-based fusion approach appears to yield the most favorable results in terms of model performance.
\begin{table}[t]
\footnotesize
\centering
\caption{Classification performance on different configurations of fusion methods.}
\label{tab:fusion}
\resizebox{0.45\textwidth}{!}{
\begin{tabular}{cllll}
\toprule
\multicolumn{1}{c}{\multirow{1}{*}{\begin{tabular}[c]{@{}l@{}}Dataset \end{tabular}}}  & \multirow{1}{*}{Metrics} &Multiplication& Addition & concatenation\\ \midrule 
\multirow{2}{*}{PhysioNet} & AUROC & \textbf{0.862 $\pm$ 0.007} & 0.860 $\pm$ 0.011 & 0.859 $\pm$ 0.008 \\ \cmidrule{2-5}
&AUPRC &{0.551 $\pm$ 0.028} & \textbf{0.553 $\pm$ 0.041}& 0.550 $\pm$ 0.040\\ \cmidrule{1-5}
\multirow{2}{*}{MIMIC IV} & AUROC & \textbf{0.817 $\pm$ 0.006} & 0.815 $\pm$ 0.004 & 0.814 $\pm$ 0.003 \\ \cmidrule{2-5}
&AUPRC &\textbf{0.335 $\pm$ 0.023} & {0.330 $\pm$ 0.013} & 0.330 $\pm$ 0.016\\ \cmidrule{1-5}
\multirow{2}{*}{Human Activity} & ACC & \textbf{0.907 $\pm$ 0.004} & {0.904 $\pm$ 0.006} & 0.902 $\pm$ 0.006\\\cmidrule{2-5}
&AUPRC & \textbf{0.906 $\pm$ 0.007} &  \textbf{0.906 $\pm$ 0.007} & 0.903 $\pm$ 0.007\\ \bottomrule
\end{tabular}}
\end{table}

\section{Conclusion}
In this paper, we presented a two-stage approach for modeling irregular multivariate time series, addressing the limitations of existing methods in tackling temporal and feature-wise data irregularities. We conducted a comprehensive comparison between our proposed method, TADA, and several state-of-the-art baselines for modeling irregular time series data. The results showed that our method outperformed all baselines on three common benchmark datasets. 


Lastly, given our proposed DLA attention module introduces additional complexity compared to the typical scaled dot-product attention, possible directions for future research would investigate exploring efficient and scalable window attention mechanisms as proposed by~\citep{swinformer,neighbohoodattention} for modeling images. Additionally, we noticed that DLA is also useful for reconstructing missing values. The next steps would be to investigate the use of our two-stage approach in an encoder-decoder structure and evaluate its performance on reconstruction tasks. Moreover, for the MIMIC IV dataset, we plan to adapt and assess our model on additional diverse tasks such as length of stay.

\subsubsection*{Acknowledgments}
This work is supported by the Swiss National Science Foundation (project 201184).

\bibliography{iclr2024_conference}

\begin{thebibliography}{37}
\providecommand{\natexlab}[1]{#1}
\providecommand{\url}[1]{\texttt{#1}}
\expandafter\ifx\csname urlstyle\endcsname\relax
  \providecommand{\doi}[1]{doi: #1}\else
  \providecommand{\doi}{doi: \begingroup \urlstyle{rm}\Url}\fi

\bibitem[Agarap(2018)]{relu}
Abien~Fred Agarap.
\newblock Deep learning using rectified linear units (relu).
\newblock \emph{arXiv preprint arXiv:1803.08375}, 2018.

\bibitem[Braun et~al.(2022)Braun, Fernandez, Eroglu, Hartland, Breitenbach, and Marwan]{braun2022sampling}
Tobias Braun, Cinthya~N Fernandez, Deniz Eroglu, Adam Hartland, Sebastian~FM Breitenbach, and Norbert Marwan.
\newblock Sampling rate-corrected analysis of irregularly sampled time series.
\newblock \emph{Physical Review E}, 105\penalty0 (2):\penalty0 024206, 2022.

\bibitem[Che et~al.(2018)Che, Purushotham, Cho, Sontag, and Liu]{GRUD}
Zhengping Che, Sanjay Purushotham, Kyunghyun Cho, David Sontag, and Yan Liu.
\newblock Recurrent neural networks for multivariate time series with missing values.
\newblock \emph{Scientific reports}, 8\penalty0 (1):\penalty0 6085, 2018.

\bibitem[Chen et~al.(2018)Chen, Rubanova, Bettencourt, and Duvenaud]{ode}
Ricky~TQ Chen, Yulia Rubanova, Jesse Bettencourt, and David~K Duvenaud.
\newblock Neural ordinary differential equations.
\newblock \emph{Advances in neural information processing systems}, 31, 2018.

\bibitem[Good(1952)]{crossentropy}
Irving~John Good.
\newblock Rational decisions.
\newblock \emph{Journal of the Royal Statistical Society: Series B (Methodological)}, 14\penalty0 (1):\penalty0 107--114, 1952.

\bibitem[Hassani et~al.(2023)Hassani, Walton, Li, Li, and Shi]{neighbohoodattention}
Ali Hassani, Steven Walton, Jiachen Li, Shen Li, and Humphrey Shi.
\newblock Neighborhood attention transformer.
\newblock In \emph{Proceedings of the IEEE/CVF Conference on Computer Vision and Pattern Recognition}, pp.\  6185--6194, 2023.

\bibitem[Hochreiter \& Schmidhuber(1997)Hochreiter and Schmidhuber]{LSTM}
Sepp Hochreiter and J{\"u}rgen Schmidhuber.
\newblock Long short-term memory.
\newblock \emph{Neural computation}, 9\penalty0 (8):\penalty0 1735--1780, 1997.

\bibitem[Hoffmann et~al.(2020)Hoffmann, Kotzur, Stolten, and Robinius]{hoffmann2020review}
Maximilian Hoffmann, Leander Kotzur, Detlef Stolten, and Martin Robinius.
\newblock A review on time series aggregation methods for energy system models.
\newblock \emph{Energies}, 13\penalty0 (3):\penalty0 641, 2020.

\bibitem[Horn et~al.(2020)Horn, Moor, Bock, Rieck, and Borgwardt]{seft}
Max Horn, Michael Moor, Christian Bock, Bastian Rieck, and Karsten Borgwardt.
\newblock Set functions for time series.
\newblock In \emph{International Conference on Machine Learning}, pp.\  4353--4363. PMLR, 2020.

\bibitem[Hornik et~al.(1989)Hornik, Stinchcombe, and White]{mlpfunction}
Kurt Hornik, Maxwell Stinchcombe, and Halbert White.
\newblock Multilayer feedforward networks are universal approximators.
\newblock \emph{Neural networks}, 2\penalty0 (5):\penalty0 359--366, 1989.

\bibitem[Johnson et~al.(2020)Johnson, Bulgarelli, Pollard, Horng, Celi, and Mark]{mimiciv}
Alistair Johnson, Lucas Bulgarelli, Tom Pollard, Steven Horng, Leo~Anthony Celi, and Roger Mark.
\newblock Mimic-iv.
\newblock \emph{PhysioNet. Available online at: https://physionet. org/content/mimiciv/1.0/(accessed August 23, 2021)}, 2020.

\bibitem[Khan et~al.(2022)Khan, Naseer, Hayat, Zamir, Khan, and Shah]{mlpreview}
Salman Khan, Muzammal Naseer, Munawar Hayat, Syed~Waqas Zamir, Fahad~Shahbaz Khan, and Mubarak Shah.
\newblock Transformers in vision: A survey.
\newblock \emph{ACM computing surveys (CSUR)}, 54\penalty0 (10s):\penalty0 1--41, 2022.

\bibitem[Kotzur et~al.(2018)Kotzur, Markewitz, Robinius, and Stolten]{kotzur2018impact}
Leander Kotzur, Peter Markewitz, Martin Robinius, and Detlef Stolten.
\newblock Impact of different time series aggregation methods on optimal energy system design.
\newblock \emph{Renewable energy}, 117:\penalty0 474--487, 2018.

\bibitem[Li \& Marlin(2020)Li and Marlin]{maskingfunction}
Steven Cheng-Xian Li and Benjamin Marlin.
\newblock Learning from irregularly-sampled time series: A missing data perspective.
\newblock In \emph{Proceedings of the 37th International Conference on Machine Learning}, volume 119 of \emph{Proceedings of Machine Learning Research}, pp.\  5937--5946. PMLR, 13--18 Jul 2020.

\bibitem[Li et~al.(2023)Li, Rao, Pan, and Xu]{mtsmixer}
Zhe Li, Zhongwen Rao, Lujia Pan, and Zenglin Xu.
\newblock Mts-mixers: Multivariate time series forecasting via factorized temporal and channel mixing.
\newblock \emph{arXiv preprint arXiv:2302.04501}, 2023.

\bibitem[Liu et~al.(2019)Liu, Hsaio, and Tu]{multiscale}
Chien-Liang Liu, Wen-Hoar Hsaio, and Yao-Chung Tu.
\newblock Time series classification with multivariate convolutional neural network.
\newblock \emph{IEEE Transactions on Industrial Electronics}, 66\penalty0 (6):\penalty0 4788--4797, 2019.
\newblock \doi{10.1109/TIE.2018.2864702}.

\bibitem[Liu et~al.(2021{\natexlab{a}})Liu, Dai, So, and Le]{attmlp}
Hanxiao Liu, Zihang Dai, David So, and Quoc~V Le.
\newblock Pay attention to mlps.
\newblock \emph{Advances in Neural Information Processing Systems}, 34:\penalty0 9204--9215, 2021{\natexlab{a}}.

\bibitem[Liu et~al.(2021{\natexlab{b}})Liu, Lin, Cao, Hu, Wei, Zhang, Lin, and Guo]{swinformer}
Ze~Liu, Yutong Lin, Yue Cao, Han Hu, Yixuan Wei, Zheng Zhang, Stephen Lin, and Baining Guo.
\newblock Swin transformer: Hierarchical vision transformer using shifted windows.
\newblock In \emph{Proceedings of the IEEE/CVF international conference on computer vision}, pp.\  10012--10022, 2021{\natexlab{b}}.

\bibitem[Marlin et~al.(2012)Marlin, Kale, Khemani, and Wetzel]{tsregular}
Benjamin~M Marlin, David~C Kale, Robinder~G Khemani, and Randall~C Wetzel.
\newblock Unsupervised pattern discovery in electronic health care data using probabilistic clustering models.
\newblock In \emph{Proceedings of the 2nd ACM SIGHIT international health informatics symposium}, pp.\  389--398, 2012.

\bibitem[Neil et~al.(2016)Neil, Pfeiffer, and Liu]{phasedLSTM}
Daniel Neil, Michael Pfeiffer, and Shih-Chii Liu.
\newblock Phased lstm: Accelerating recurrent network training for long or event-based sequences.
\newblock \emph{Advances in neural information processing systems}, 29, 2016.

\bibitem[Nerlove et~al.(2014)Nerlove, Grether, and Carvalho]{nerlove2014analysis}
Marc Nerlove, David~M Grether, and Jose~L Carvalho.
\newblock \emph{Analysis of economic time series: a synthesis}.
\newblock Academic Press, 2014.

\bibitem[Rubanova et~al.(2019)Rubanova, Chen, and Duvenaud]{latenode}
Yulia Rubanova, Ricky~TQ Chen, and David~K Duvenaud.
\newblock Latent ordinary differential equations for irregularly-sampled time series.
\newblock \emph{Advances in neural information processing systems}, 32, 2019.

\bibitem[Shukla \& Marlin(2019)Shukla and Marlin]{ipnet}
Satya~Narayan Shukla and Benjamin~M Marlin.
\newblock Interpolation-prediction networks for irregularly sampled time series.
\newblock \emph{arXiv preprint arXiv:1909.07782}, 2019.

\bibitem[Shukla \& Marlin(2021)Shukla and Marlin]{mtan}
Satya~Narayan Shukla and Benjamin~M Marlin.
\newblock Multi-time attention networks for irregularly sampled time series.
\newblock \emph{arXiv preprint arXiv:2101.10318}, 2021.

\bibitem[Silva et~al.(2012)Silva, Moody, Scott, Celi, and Mark]{physionet2012}
Ikaro Silva, George Moody, Daniel~J Scott, Leo~A Celi, and Roger~G Mark.
\newblock Predicting in-hospital mortality of icu patients: The physionet/computing in cardiology challenge 2012.
\newblock In \emph{2012 Computing in Cardiology}, pp.\  245--248. IEEE, 2012.

\bibitem[Sun et~al.(2020)Sun, Hong, Song, and Li]{samplingratereview}
Chenxi Sun, Shenda Hong, Moxian Song, and Hongyan Li.
\newblock A review of deep learning methods for irregularly sampled medical time series data.
\newblock \emph{arXiv preprint arXiv:2010.12493}, 2020.

\bibitem[Teichgraeber \& Brandt(2022)Teichgraeber and Brandt]{teichgraeber2022time}
Holger Teichgraeber and Adam~R Brandt.
\newblock Time-series aggregation for the optimization of energy systems: Goals, challenges, approaches, and opportunities.
\newblock \emph{Renewable and Sustainable Energy Reviews}, 157:\penalty0 111984, 2022.

\bibitem[Tolstikhin et~al.(2021)Tolstikhin, Houlsby, Kolesnikov, Beyer, Zhai, Unterthiner, Yung, Steiner, Keysers, Uszkoreit, et~al.]{mixer}
Ilya~O Tolstikhin, Neil Houlsby, Alexander Kolesnikov, Lucas Beyer, Xiaohua Zhai, Thomas Unterthiner, Jessica Yung, Andreas Steiner, Daniel Keysers, Jakob Uszkoreit, et~al.
\newblock Mlp-mixer: An all-mlp architecture for vision.
\newblock \emph{Advances in neural information processing systems}, 34:\penalty0 24261--24272, 2021.

\bibitem[Vaswani et~al.(2017{\natexlab{a}})Vaswani, Shazeer, Parmar, Uszkoreit, Jones, Gomez, Kaiser, and Polosukhin]{vaswani2017attention}
Ashish Vaswani, Noam Shazeer, Niki Parmar, Jakob Uszkoreit, Llion Jones, Aidan~N Gomez, \L~ukasz Kaiser, and Illia Polosukhin.
\newblock Attention is all you need.
\newblock In \emph{Advances in Neural Information Processing Systems}, volume~30, 2017{\natexlab{a}}.
\newblock URL \url{https://proceedings.neurips.cc/paper_files/paper/2017/file/3f5ee243547dee91fbd053c1c4a845aa-Paper.pdf}.

\bibitem[Vaswani et~al.(2017{\natexlab{b}})Vaswani, Shazeer, Parmar, Uszkoreit, Jones, Gomez, Kaiser, and Polosukhin]{transformer}
Ashish Vaswani, Noam Shazeer, Niki Parmar, Jakob Uszkoreit, Llion Jones, Aidan~N Gomez, {\L}ukasz Kaiser, and Illia Polosukhin.
\newblock Attention is all you need.
\newblock \emph{Advances in neural information processing systems}, 30, 2017{\natexlab{b}}.

\bibitem[Weerakody et~al.(2021)Weerakody, Wong, Wang, and Ela]{weerakody2021review}
Philip~B Weerakody, Kok~Wai Wong, Guanjin Wang, and Wendell Ela.
\newblock A review of irregular time series data handling with gated recurrent neural networks.
\newblock \emph{Neurocomputing}, 441:\penalty0 161--178, 2021.

\bibitem[Zaheer et~al.(2017)Zaheer, Kottur, Ravanbakhsh, Poczos, Salakhutdinov, and Smola]{deepset}
Manzil Zaheer, Satwik Kottur, Siamak Ravanbakhsh, Barnabas Poczos, Russ~R Salakhutdinov, and Alexander~J Smola.
\newblock Deep sets.
\newblock \emph{Advances in neural information processing systems}, 30, 2017.

\bibitem[Zeng et~al.(2023)Zeng, Chen, Zhang, and Xu]{linearts}
Ailing Zeng, Muxi Chen, Lei Zhang, and Qiang Xu.
\newblock Are transformers effective for time series forecasting?
\newblock In \emph{Proceedings of the AAAI conference on artificial intelligence}, pp.\  11121--11128, 2023.

\bibitem[Zhang et~al.(2023)Zhang, Zheng, Cao, Bian, and Li]{warpformer}
Jiawen Zhang, Shun Zheng, Wei Cao, Jiang Bian, and Jia Li.
\newblock Warpformer: A multi-scale modeling approach for irregular clinical time series.
\newblock \emph{arXiv preprint arXiv:2306.09368}, 2023.

\bibitem[Zhang et~al.(2021)Zhang, Zeman, Tsiligkaridis, and Zitnik]{raindrop}
Xiang Zhang, Marko Zeman, Theodoros Tsiligkaridis, and Marinka Zitnik.
\newblock Graph-guided network for irregularly sampled multivariate time series.
\newblock \emph{arXiv preprint arXiv:2110.05357}, 2021.

\bibitem[Zhang \& Yan(2022)Zhang and Yan]{crossformer}
Yunhao Zhang and Junchi Yan.
\newblock Crossformer: Transformer utilizing cross-dimension dependency for multivariate time series forecasting.
\newblock In \emph{The Eleventh International Conference on Learning Representations}, 2022.

\bibitem[Zhou et~al.(2022)Zhou, Ma, Wen, Wang, Sun, and Jin]{fedformer}
Tian Zhou, Ziqing Ma, Qingsong Wen, Xue Wang, Liang Sun, and Rong Jin.
\newblock Fedformer: Frequency enhanced decomposed transformer for long-term series forecasting.
\newblock In \emph{International Conference on Machine Learning}, pp.\  27268--27286. PMLR, 2022.

\end{thebibliography}
\bibliographystyle{iclr2024_conference}

\appendix
\section{Related Work: Multi-layer Perceptron Mixer}
The Transformer~\citep{vaswani2017attention} has demonstrated its effectiveness in various domains such as natural language processing (NLP), computer vision (CV), and time series analysis. The Transformer architecture with a multi-head self-attention mechanism achieved competitive results when dealing with long sequences. However, there have been studies investigating the necessity of the self-attention blocks, leading to the proposal of MLP-based alternatives~\citep{attmlp,mixer}.  For example, the MLP-Mixer proposed in~\citep{mixer}  performs patch and channel mixing without self-attention blocks applied on CV tasks,  where images are divided into distinct non-overlapping patches analogous to tokens in NLP. Similarly, the authors in \cite{attmlp} proposed an MLP-based structure with gating to improve the cross-token (i.e., patch) interactions. These simpler architectures have shown promising results in vision-related tasks, suggesting that MLPs could efficiently represent any function with a fixed parameterization~\citep{mlpfunction}, and attention blocks in transformers may not be the primary determinant of excellent performance. Instead, the mixers that alternatively mix along different dimensions, for instance, in multi-variate time series, by first mixing along the feature dimension and then mixing along the time dimension, could offer strong competition to transformers~\citep{mlpreview}. 
In time series forecasting, for example, the work in \cite{linearts} demonstrates that a simple linear layer beats transformer-based models. Similarly, MTSMixer~\citep{mtsmixer} illustrates that MLP is sufficient for capturing temporal and feature-based interactions. Nonetheless, there is still a lack of research into using MLP mixers for time series classification and more so for dealing with irregular signals. 

\section{Datasets}\label{appendix:data}
The \textbf{PhysioNet Challenge 2012} dataset~\citep{physionet2012} comprises 12,000 individual stays in the ICU, each lasting a minimum of 48 hours. Each ICU stay is represented as a multivariate time series, including up to 37 variables and 4 types of demographic information. It's important to note that the length of the time series and the number of features can vary for each patient. We adopted the processing methods described by ~\citep{latenode}. The label of each patient is assigned based on patient mortality during their ICU stay. We divided the data into training, validation, and testing sets in an 8:1:1 ratio. Given the high imbalance in the dataset, with only 13.8\% of positive labels, we preserved similar label proportions in all data partitions. The primary task for this dataset is to predict in-hospital mortality.

The \textbf{MIMIC-IV} dataset~\citep{mimiciv} is a multivariate time series dataset consisting of sparse and irregularly sampled physiological signals. MIMIC-IV is an updated version of the MIMIC-III dataset with a more significant number of patients. It offers an extensive clinical data repository from over 400,000 patients admitted to the Beth Israel Deaconess Medical Center between 2008 and 2019. Following a similar procedure in~\citep{ipnet}, we extracted 12 standard physiological variables from 59,843 samples after removing stays shorter than 48 hours. Similar to the PhysioNet dataset, the length of the time series and the number of features can vary for each patient. We trained and evaluated our model on the binary mortality prediction task. We divided the data into training, validation, and testing sets in a 6:2:2 ratio. Given the high imbalance in the dataset, with only 8.7\% of positive labels, we preserved similar label proportions in all partitions.

 The \textbf{Human Activity} dataset~\citep{latenode} consists of 6,554 samples, each containing up to 12 variables. All instances in the dataset have a fixed length of 50 timesteps. We followed the steps in \cite{mtan} to process and split the dataset. For each sample, each timestep is classified into one of eleven activity categories, such as walking, standing, and sitting. Our model was trained and evaluated on the task of classifying each timestep in the sequence.

 To feed the data into neural networks, we set the input as zero if no value was measured and normalized all time values within each sample. The dataset statistics, including missing ratio, are provided in Table~\ref{tab:datasets}
\begin{table*}[tb!]
\centering
\caption{Dataset statistics. 'Avg. length' indicates the average number of observed timestamps present in the dataset. The `classes' means the number of categories in dataset labels. The `missing' denotes the missing ratio between the number of unobserved features and the number of all possible observations. For datasets with binary labels, "Positive Label" refers to the ratio of positive labels in the dataset.}
\label{tab:datasets}
\begin{tabular}{lcccccc}
\toprule
Datasets & \multicolumn{1}{l}{size} & \multicolumn{1}{c}{features} & \multicolumn{1}{c}{Avg. length} & \multicolumn{1}{c}{classes} & \multicolumn{1}{c}{missing (\%)}& \multicolumn{1}{c}{positive (\%)} \\ 
\midrule
PhysioNet & 12,000 & 41 & 60 & 2  & 88.4 & 14.2\\
MIMIC IV & 59,843 & 12 & 70 & 2  & 72.8 & 8.7\\
Human Activity & 6,554 & 12 & 50 & 7  & 75.0 & -\\ 
\bottomrule
\end{tabular}
\end{table*}

\section{Details of our Model}
\subsection{Hierarchical MLP Mixers}

\begin{wrapfigure}{r}{0.45\textwidth}
  \vspace{-3mm}
   \centering
   \begin{minipage}{0.45\textwidth}
    \centering
    \includegraphics[width=\textwidth]{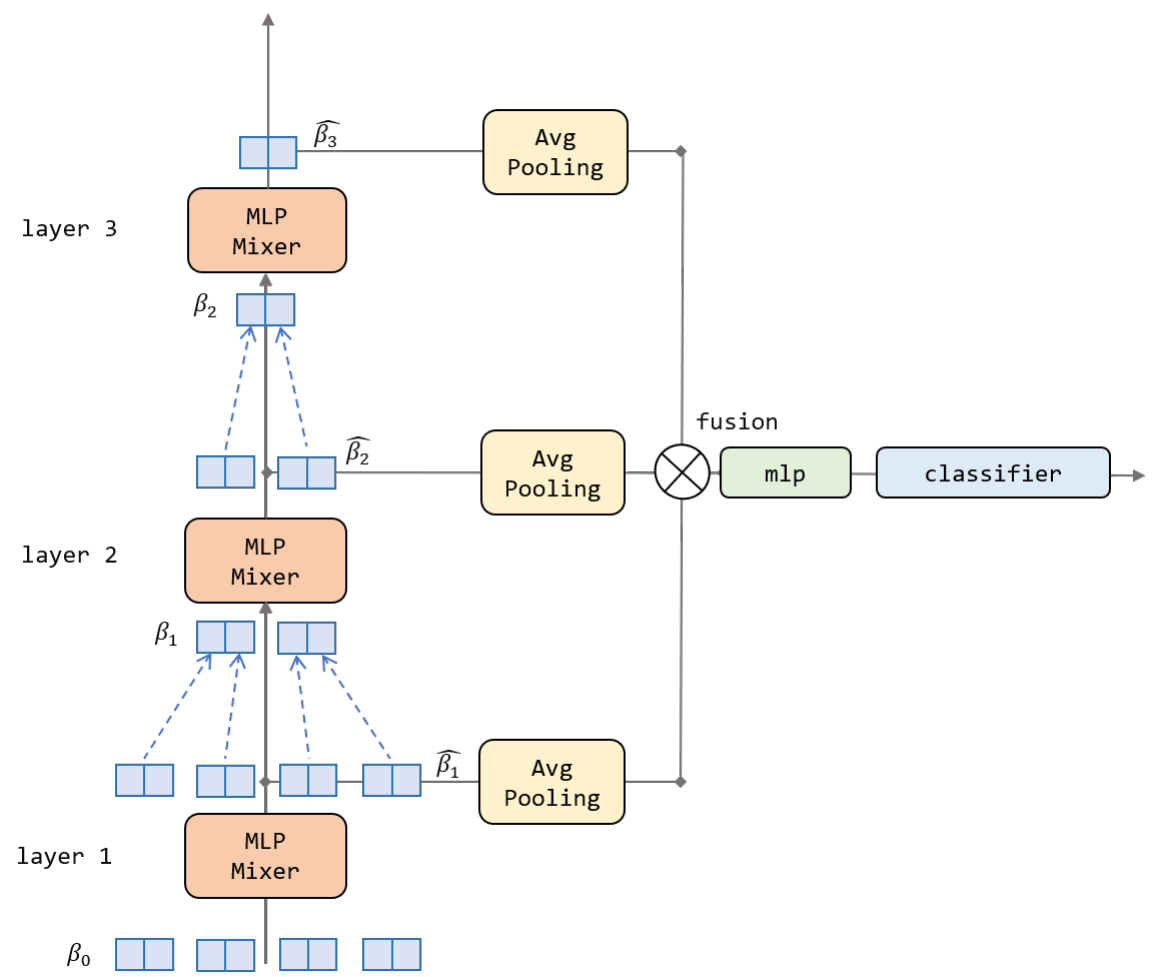}
   \caption{Illustration of the hierarchical structure and the fusion block.}
   \label{fig:fusion}
 \end{minipage}
 \vspace{-3mm}
\end{wrapfigure}

The hierarchical MLP mixers consist of a series of blocks that share the same architecture. After processing with \texttt{DLA}, we obtain an output $A\in\mathbb{R}^{L \times d}$ where $L$ represents the time dimension (proxy to time dimension) and $d$ represents the feature dimension. We patch the matrix $A$ across the time dimension into  patches $\beta^0_{i,:}\in\mathbb{R}^{p\times d}$, where $1\leq i\leq L/p$. For convenience, we concatenate all patches together as $\beta^0 = \text{concat}[\beta^0_{1,:},\beta^0_{2,:},...\beta^0_{L_0,:}]$, where $\beta^0 \in \mathbb{R}^{L_0 \times p \times d}$, $L_0 = L/p$. 
The concatenation of patches are inputs of each MLP mixer block. For $l\geq 1$, each block is defined as: 
\begin{equation}\label{eqappendix: mlpblock}
\begin{gathered}
    \hat{\beta}^{l} = \sigma(\beta^{l-1}+(W_2\beta^{l-1}W_1)W_3)
\end{gathered}
\end{equation}
\begin{equation}\label{eq:supp patch merge}
\begin{aligned}
    \beta^{l}_{i,:} &= \text{concat}\left [ W_4 \hat{\beta}^{l}_{mi-1,:},...,W_4 \hat{\beta}^{l}_{mi-(m-1),:} \right ]
\end{aligned}
\end{equation}
where $W_1,W_3\in\mathbb{R}^{d\times d}$, $W_2\in\mathbb{R}^{L_{l-1}\times L_{l-1}}$,$W_4\in\mathbb{R}^{p\times p/m}$. $\beta_{i,:}^l \in\mathbb{R}^{p\times d}$ denotes the $i$th patch in $\beta^l$, constrained by $1\leq i\leq \frac{L_{l}}{p/m}$. Each patch of length $p$ is transformed to a length of $p/m$ through $W_4$. The pseudocode of each block is shown in code listing~\ref{lst:mlp mixer}, where \texttt{Linear} denotes a linear projection along certain dimension of the input vector.

\begin{listing}[tb]%
\caption{pseudocode for MLP mixer}%
\label{lst:mlp mixer}%
\begin{lstlisting}[language=python,mathescape]
def mlp_mixer(x, d_model,p_num,p_size, m):
  shortcut = x 
  x = Linear(x, d_model, axis= 2) \\ $W_1$
  x = Linear(x, p_num, axis= 0) \\ $W_2$
  x = Linear(x, d_model, axis= 2) \\ $W_3$
  x_out = relu(x + shortcut)
  x = Linear(x_out, p_size/m,axis = 1)\\$W_4$
  x = reshape('p_num, p_size/m, d_model'-> 'p_num/m, p_size, d_model')
  return x,x_out
\end{lstlisting}
\end{listing}

Each MLP mixer block produces two outputs, one for the classifier and one for the next block.

To extract different scales of information, we fuse the results from the different blocks by a fusion block as illustrated in Figure~\ref{fig:fusion}. 
The average pooling layer projects each layer's output to the same vector dimension. Then we use an element-wise product and an MLP layer to fuse the different scales of $\hat{\beta}$. The final output is passed through a classifier. 

\subsection{Classifier and Loss function}
Given the final vector representation $c\in\mathbb{R}^{l_c\times d_c}$, we use a simple classifier for the outcome prediction. This representation initially passes through an average pooling layer to modify its length. For Physionet and MIMIC IV datasets, we pool the size to $\mathbb{R}^{1\times d_c}$. In the case of Human Activity dataset, which requires a label prediction at each time step, the size of $c$ is pooled to $\mathbb{R}^{l_{\text{act}}\times d_c}$, where $l_{\text{act}}$ is the length of the sample in Human Activity dataset. After pooling, a linear layer is applied to project $c$ to either $\mathbb{R}^{1\times n_{\text{class}}}$ or $\mathbb{R}^{l_{\text{act}}\times n_{\text{class}}}$, depending on the dataset used. $n_{class}$ denotes the number of classes in that dataset. The loss function is the cross-entropy loss~\citep{crossentropy}.
\section{Transformer and MLP mixers}
\begin{table}[ht]
\footnotesize
\centering
\caption{Classification performance on samples MLP mixers or transformer encoder. }
\label{tab:transformer}
\begin{tabular}{c|l|ll}
\toprule
\multicolumn{1}{c}{\multirow{1}{*}{\begin{tabular}[c]{@{}l@{}}Dataset \end{tabular}}}  & \multirow{1}{*}{Metrics} &MLP Mixer& Transformer\\ \midrule 
\multirow{2}{*}{PhysioNet} & AUROC & \textbf{0.862 $\pm$ 0.007} & 0.857 $\pm$ 0.010 \\ \cmidrule{2-4}
&AUPRC &\textbf{0.551 $\pm$ 0.028} & 0.549 $\pm$ 0.029\\ \cmidrule{1-4}
\multirow{2}{*}{MIMIC IV} & AUROC & \textbf{0.817 $\pm$ 0.006} & 0.811 $\pm$ 0.002 \\ \cmidrule{2-4}
&AUPRC &\textbf{0.335 $\pm$ 0.023} & {0.328 $\pm$ 0.010}\\ \cmidrule{1-4}
\multirow{2}{*}{Human Activity} & Accuracy & \textbf{0.907 $\pm$ 0.004} & {0.893 $\pm$ 0.012} \\\cmidrule{2-4}
&AUPRC & \textbf{0.906 $\pm$ 0.007} &  0.872 $\pm$ 0.024  \\ \bottomrule
\end{tabular}
\end{table}
We conducted a comparison between our proposed method and a transformer-based approach, given that the MLP Mixer was introduced as a more straightforward alternative to self-attention~\citep{mixer}. Specifically, we evaluated the model by substituting the MLP Mixer block with a transformer-based block. This is done by replacing the linear patch pooling layer with a scaled dot-product attention operation.

For simplicity, we define the scaled dot-product attention block as:
\begin{equation}\label{eq:attn block}
\begin{aligned}
    \texttt{AttnBlock}(X,Y,Y) &= M+\texttt{Feedforward}(M),\\
    M &= X+\texttt{MSA}(X,Y,Y)
\end{aligned}
\end{equation}
where \texttt{MSA} denotes the multi-head self-attention proposed in \cite{transformer}, \texttt{Feedfoward} denotes the MLP layer.

Let the input of the $l$-th block be $\beta^l\in\mathbb{R}^{L_l\times p\times d}$, $\beta^l_{:,j}$ denote all time steps in dimension $j$. The transformer-based block is defined as:
\begin{equation}\label{eq:transformer}
\begin{aligned}
    \bar{\beta}^l_{:,j} = \texttt{AttnBlock}(\beta^{l-1}_{:,j},\beta^{l-1}_{:,j},\beta^{l-1}_{:,j})\\
\end{aligned}
\end{equation}
where $1\leq j\leq p$. The linear pooling layer ($W_4$ in Equation~\ref{eq:supp patch merge}) is replaced by an attention operation:
\begin{equation}\label{eq:attn pool}
\begin{aligned}
    \hat{\beta}^l_{i,:} = \texttt{AttnBlock}(Q_{i,:},\bar{\beta}^{l}_{i,:},\bar{\beta}^{l}_{i,:}),
    1\leq i\leq L_l
\end{aligned}
\end{equation}
where $Q\in\mathbb{R}^{L_l\times p/m\times d}$ is the learnable query matrix responsible for pooling patch of size $p$ to size $p/m$ by in Equation~\ref{eq:attn pool}. Subsequently, the patches with reduced size will be regrouped into new patches: 
\begin{equation}
\begin{aligned}
    \beta^{l}_{i,:} &= \text{concat}\left [\hat{\beta}^{l}_{mi-1,:},...,\hat{\beta}^{l}_{mi-(m-1),:} \right ], 1\leq i\leq L_l/m
\end{aligned}
\end{equation}
The output $\hat{\beta}^l\in\mathbb{R}^{L_l\times p/m\times d}$ represents the sequence that contains all the patches after pooling. 

The performance of the two representation learning methods is presented in Table~\ref{tab:transformer}. Our experiments demonstrate that with previous \texttt{DLA}, leveraging simple linear features is sufficient for downstream classification.

\section{Temporal Embeddings}
\begin{table}[!ht]
\centering
\footnotesize
\caption{Classification performance on different configurations of constructing keys and values. 'Ours' designates our proposed model, utilizing the mask function for values and TE function for keys. 'Setting (1)' refers to applying the same mask function to both values and keys. Setting (2) refers to apply the same TE function to both values and keys.}
\label{tab:embedablation}
\resizebox{0.45\textwidth}{!}{
\begin{tabular}{c|l|lll}
\toprule
\multicolumn{1}{c}{\multirow{1}{*}{\begin{tabular}[c]{@{}l@{}}Dataset \end{tabular}}}  & \multirow{1}{*}{Metrics} &Ours& Setting (1) & Setting (2)\\ \midrule 
\multirow{2}{*}{PhysioNet} & AUROC & \textbf{0.862 $\pm$ 0.007} & 0.857 $\pm$ 0.003 & 0.588 $\pm$ 0.022 \\ \cmidrule{2-5}
&AUPRC &\textbf{0.551 $\pm$ 0.028} & 0.541 $\pm$ 0.032& 0.198 $\pm$ 0.016\\ \cmidrule{1-5}
\multirow{2}{*}{MIMIC IV} & AUROC & \textbf{0.817 $\pm$ 0.006} & 0.816 $\pm$ 0.010 & 0.660 $\pm$ 0.003 \\ \cmidrule{2-5}
&AUPRC &0.335 $\pm$ 0.023 & \textbf{0.336 $\pm$ 0.010} & 0.139 $\pm$ 0.004\\ \cmidrule{1-5}
\multirow{2}{*}{Human Activity} & Accuracy & \textbf{0.907 $\pm$ 0.004} & {0.906 $\pm$ 0.010} & 0.380 $\pm$ 0.008\\\cmidrule{2-5}
&AUPRC & \textbf{0.906 $\pm$ 0.007} &  0.884 $\pm$ 0.013 & 0.143 $\pm$ 0.001\\ \bottomrule
\end{tabular}}
\end{table}
Furthermore, to demonstrate the importance of the attention-based temporal embedding and the choice of values and keys, we conducted additional model configurations as follows: (1) Employing the same mask function for processing both DLA values and keys, and (2) Utilizing an identical temporal embedding for encoding both DLA values and keys. Table~\ref{tab:embedablation} shows the results of different settings. Our observations highlight the critical importance of retaining original input data as values, as evident from the substantial drop in performance observed in setting (2).  Moreover, in the context of setting (1), we note that within the PhysioNet dataset, leveraging temporal embedding yields performance improvements. highlighting the importance of applying TE to learn dimension correlations.

\end{document}